\documentclass[12pt]{article}
\usepackage[english]{babel}
\usepackage{amsmath}
\usepackage{graphicx}
\usepackage{textcomp}
\usepackage{parskip}
\usepackage[colorinlistoftodos]{todonotes}
\usepackage{csquotes}
\usepackage{float}
\usepackage{comment}
\usepackage{caption}
\usepackage[numbers]{natbib}
\usepackage{pdfpages}
\usepackage{fancyhdr}
\usepackage{multirow}
\usepackage{url}

\title{Defining the problem of Observation Learning }
\author{Leo Pauly\\
cnlp@leeds.ac.uk\\
University of Leeds, UK}
\date{}

\begin{document}
\maketitle

\section*{Introduction}
\setlength{\parindent}{5ex}
\hspace{5ex}
The 21st century has seen a considerable growth in robotics technologies. More and more advanced robots are being developed every day. From room cleaning robot Roomba~\cite{forlizzi2006service} to advanced medical robot DaVinci~\cite{leven2005davinci}, robots are making way into human lives. It is widely expected that the robots will replace humans in many of the daily labor intensive and repetitive tasks in a very near future.

But one of the major challenges faced by even the most advanced robotic technologies lies in its inability to learn new tasks and skills like human beings, i.e from experiences and from demonstrations. The traditional method of teaching robots to perform new tasks is by programming each aspect of the task line by line. But it is not practical to program each and every task in such an detailed manner.The real solution for this problem lies in developing learning technologies for robots that could learn and then act just like human beings i.e. from examples and demonstrations. A 3 year old kid is taught something new by its parents not by algorithmic programming but rather by simple demonstrations. The kid is shown what to do by its parents and it learns to imitate from these behaviors.

We term this as \textbf{observation learning problem}. In observation learning, the robot is made to observe a demonstrator performing a particular task. It is then made to learn to perform this task from these observations only. The novelty of this approach lies in the learning process where the robot learns a new task by only observing a demonstration without being explicitly programmed, there by providing a robot with human like learning capabilities. A detailed formulation of the problem of observation learning is given in the next two sections.

Observation learning has close similarities to the imitation learning problem (also called as apprenticeship learning/ programming by demonstration)~\cite{PastorLearningDemonstration}~\cite{Rahmatizadeh2017Vision-BasedDemonstration} \cite{lesort2017unsupervised} which has been studied extensively using techniques like Inverse reinforcement learning \cite{abbeel2004apprenticeship} and Behavioral cloning \cite{bojarski2016end}. It can be seen in literature that these two problems are not well defined from each other separately and the terms are even used interchangeably. But observation learning clearly differs from imitation learning in the aspect that in imitation learning, the learner (the robot) can observe the demonstrator in an ego centric (first person) view, whereas in observation learning the learner can only observe the demonstrator in allocentric (third person) view. Hence the problem of observation learning has to be treated separately from imitation learning and separate tools and techniques have to be developed for solving it efficiently. In the literature, observation learning is also referred to as third person imitation learning \cite{stadie2017third} and imitation from observation \cite{Liu2017ImitationTranslation}. But in this article we coin this new term \textbf{observation learning} to differentiate it from imitation learning and indicate the unique nature of this problem. In the following sections we explain in detail the problem of observation learning, along with its mathematical formulation and literature survey.

\setlength{\parindent}{5ex}

\section*{Problem description}
\begin{figure}[h]
\includegraphics[width=15cm,height=11cm,keepaspectratio]{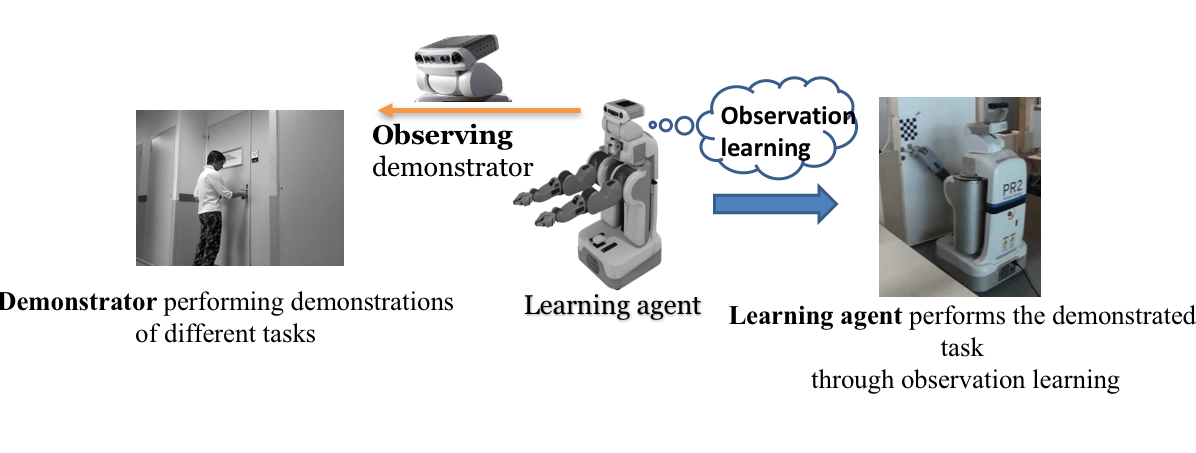}
\caption{Observation learning}
\label{fig:2.1}
\end{figure}

\hspace{5ex}Observation learning could be defined as the process of learning to perform new tasks by visually observing the demonstrations by a demonstrator performing the same task. It aims at mimicking the learning process by which humans learn new skills.
The general set up of an observation learning setting is show in the Fig \ref{fig:2.1}.

The observation learning problem has 2 objects and 2 process involved.
The two objects involved are the demonstrator and the learning agent:

\begin{itemize}
\item \textbf{Demonstrator}: It is the object that performs the demonstration of the activity/task.

\item \textbf{Learning agent}: Learning agent is the object that learns a particular task from observing the demonstrator.
\end{itemize}
The two processes involved are the observation and learning processes:
\begin{itemize}
\item \textbf{Observation}: This is the process by which the learning agent views the demonstrator. It is desirable that the learning agent does not capture any predefined features like hand motion of the demonstrator but rather captures only the raw image/video input of the demonstrator. This will help to model the  problem of observation learning much more similar to the human learning process where no explicit feature tracking is used.

\item \textbf{Learning}: In this process the learning agents learns from the demonstrations it has observed from the demonstrator. Once learned, the learning agent then performs the same action. The learning process here primarily involves the process of understanding the mapping from the observations made by the learning agent to the control variables of the manipulator of the robotic system that are required to replicate the task observed. The theories about human learning process from cognitive science and neuroscience could be incorporated in this stage for designing advanced learning processes for the learning agent.
\end{itemize}

\section*{Mathematical formulation}

\hspace{5ex}Let the set $X=\{x_{1},...,x_{n}\}$ represent a video sequence of the demonstration example performed by a human, where $x_{i}$ denotes the $i^{th}$ video frame and $n$ denotes the length of the video sequence. Let $C=\{c_{1},...,c_{t}\}$ be the control variables vector representing the joint configuration, torsion, speed and acceleration  of the robotic manipulator that determines the movements of the  robotic manipulator.

The solution to the observation learning problem has to learn a mapping $M: X\rightarrow C$ where $X$ is the video demonstration by the human demonstrator and $C$ is the control variables of the robotic arm.

Since a direct mapping is difficult to learn it would be desirable to first learn a mapping to an intermediate features space $F$ which would be a view point invariant feature space and then map from that features space $F$ to the robotic control variable feature space $C$.

So the original mapping $M$ could be split into two: $M_{1}$ and $M_{2}$.Mapping $M_{1}$ will first create a mapping to a view point invariant space $F$ from $X$. The specialty of the feature space $F$ is that, each point in that feature space represent a particular task independent of the view point. So if $X^{(1)}$ \& $X^{(2)}$ represent the video of an action $A$ taken from two viewpoints, the mapping $M_{1}$ should map both $X^{(1)}$ \& $X^{(2)}$  to the same point in $F$, i.e $M_{1}(X^{(1)})= M_{1}(X^{(2)})$.

Once this feature representation in $F$ is obtained for an action then mapping $M_{2}$  will map this point representing an action to the control variables space $C$ of the robotic manipulator, giving the joint configurations, speed, torsion and acceleration required by the manipulator for performing the same action.

Thus by learning these two mappings M1 and M2 the system effectively learns to perform an action just by observing the demonstrator. The Fig~\ref{fig:2.2} shows the pictorial illustration of this mappings :

\begin{figure}[H]
\includegraphics[width=14cm,height=8cm]{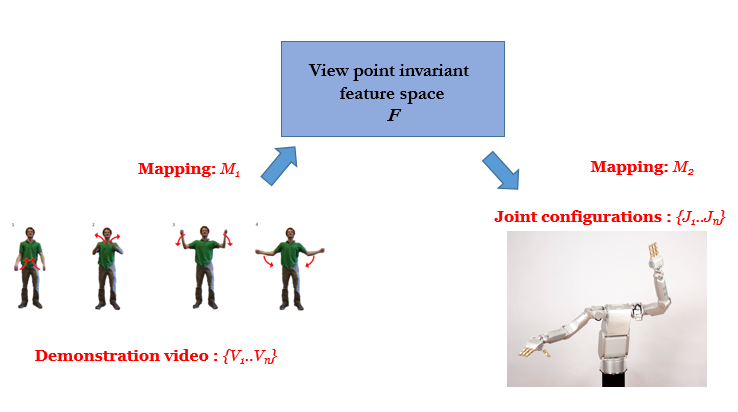}
\caption{Mathematical representation : In a nutshell}
\label{fig:2.2}
\end{figure}

\section*{Literature survey}
\setlength{\parindent}{5ex}

\hspace{5ex}This section presents a summary of the existing methods for tackling the observation learning problem. As described in the introduction, the observation learning problem has a considerable difference from the problem of imitation learning and hence is not included in this literature survey. In general observation learning can be divided into two categories:
\begin{itemize}
\item Implicit observation learning
\item Assisted observation learning
\end{itemize}

\subsection*{Implicit observation learning}
\hspace{5ex}Implicit observation learning is the process of observation learning from the raw visuals of the demonstrations performed by the demonstrator. The learning algorithm employed by the learning agent will be responsible for extracting the required information/features from the visuals of the demonstrations to learn the particular task being demonstrated.

~\citet{stadie2017third} presents a generative adversarial network (GAN)~\cite{goodfellow2014generative} based approach for tackling the implicit observation learning problem. This method is an extension to the work by~\citet{ho2016generative} to third person settings. The proposed method casts the observation learning problem as an inverse reinforcement learning (IRL) problem and uses the basic idea of GANs to solve this IRL problem.

In this paper, a GAN based method is used for inferring the reward function which is then used for training the learning agent.  The discriminator in the GAN settings tries to distinguish between the two pair of observations received by the learning agent itself. These two pairs of observations are: 3rd person observation of the actions performed by the demonstrator and the ego centric  observations of the actions performed by the learning agent. The inverse of the loss of the discriminator will serve as the reward to the reinforcement learning employed in the learning agent, which will try to reduce this lose function (there by maximizing the reward). As this loss function reduced the difference between the actions made by the learning agent and demonstrator reduces there by the learning agent acquiring the desired behavior.

\begin{figure}[H]
\centering
\includegraphics[width=17cm,height=14cm,keepaspectratio]{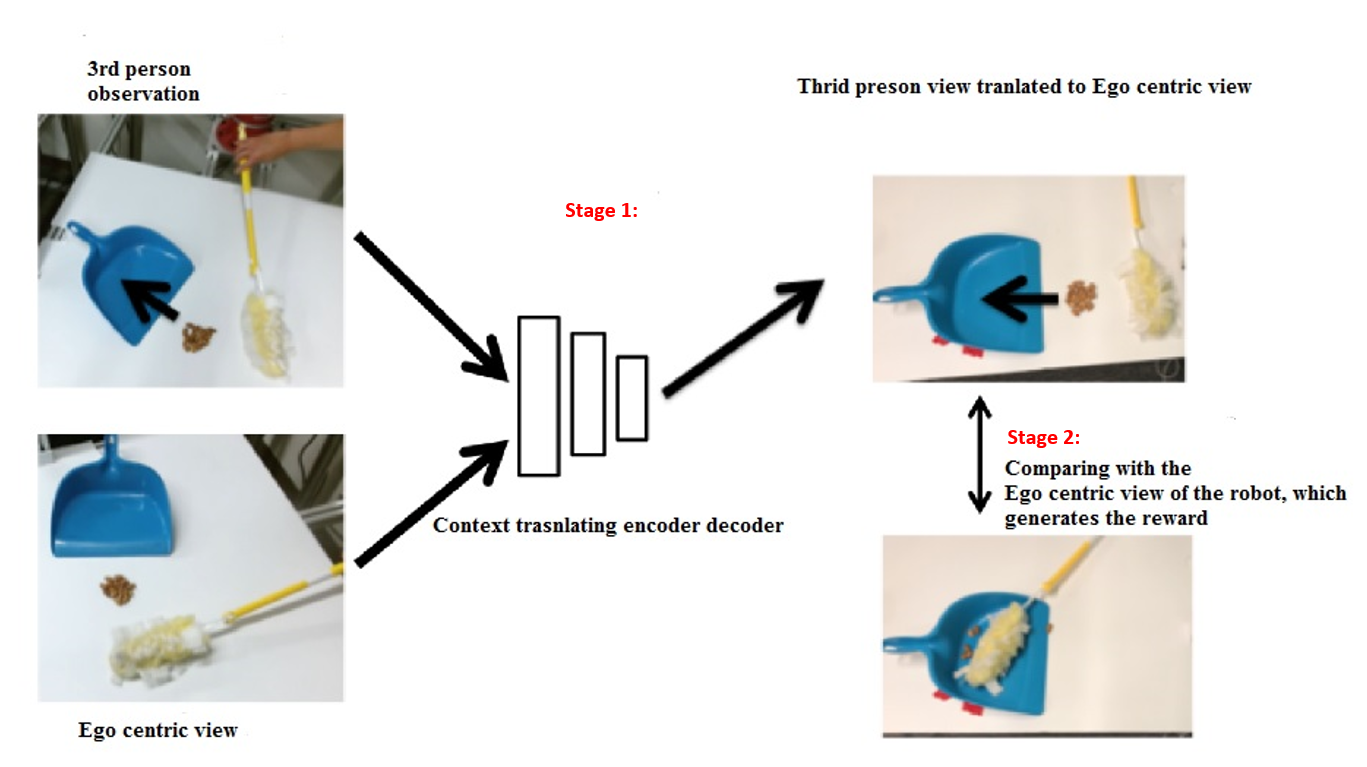}
\caption{ Observation learning using context translation}
\label{fig:app1.1}
\end{figure}

A similar approach is also presented by ~\citet{Liu2017ImitationTranslation} for tackling the problem of implicit observation learning. Here the observation leaning problem is again treated as an inverse reinforcement learning problem where 3rd person observation of the demonstrator is given and the learning agent has to infer a reward function from these observations. Here a reward function is extracted by using a process of context translation. The framework of context translation is shown in Fig \ref{fig:app1.1}.

The proposed method can be divided into to 3 stages:

\begin{enumerate}
\item Context  translation: This is the first stage.  Here the 3rd person observation of the demonstrator made by the learning agent is converted into the ego centric view of the learning agent using a encoder- decoder network. It is marked as stage 1 in Fig \ref{fig:app1.1}.
\item Obtaining reward functions: In this stage the reward function is inferred. The ego centric view of the learning agent performing the action  is compared with that of the context translated view of the demonstrator. The similarity between the features extracted from these two images is then calculated and is taken as the reward function. The same decoder in context translation is used for extracting the features for calculating the similarity between the images. The hypothesis behind this is that as the as the learning agent tries to increase the similarity of its actions with respect to that of the demonstrator, it will learn to perform action that are similar to that of the demonstrator and there by learning the demonstrated task.
\item Reinforcement learning using the reward function obtained:
Once the reward function is extracted, it is then used for training the reinforcement learning algorithm used by the learning agent. The RL algorithm will find a policy that will maximize the reward that was obtained in stage 2.
\end{enumerate}

But these methods have several drawbacks. The GAN based method~\cite{Stadie2017Third-PersonLearning} is only studied in simulated RL experiments. There is a large domain shift between the simulated world and the real world. Similarly the context translation method~\cite{Liu2017ImitationTranslation} also suffers the credibility in real world conditions. Even though few real world experiments are presented in the paper, those experiments are conducted in highly constrained environments. Also it requires large number of demonstrations for learning a single task, which is difficult to collect. Again in general it is difficult to get the reinforcement learning working in situations where high dimensional inputs like raw images are fed as observations. A possible solutions to this is to use non-RL based methods.

\begin{figure}[H]
\centering
\includegraphics[width=.6\linewidth]{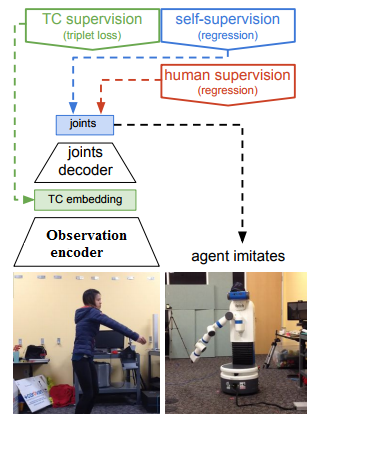}
\caption{ General framework of the proposed method of observation learning}
\label{fig:app1.2}
\end{figure}

~\citet{Sermanet2017Time-ContrastiveObservation} presents a totally new approach for solving the problem of implicit observation learning. This method does not use any reinforcement learning like the GAN based method~\cite{stadie2017third} or the context translation method~\cite{Liu2017ImitationTranslation}, rather uses an auto encoder method to learn the direct mapping from observation of the human demonstrator to the joint angles of the robotic manipulator. The Fig \ref{fig:app1.2} shows the general frame work of the proposed method.

The observation encoder encodes the raw RGB observations made by the system into a feature vector in the feature space. The joints decoder decodes this embedding to joint configurations which could be fed to the learning agent.

\noindent The proposed method employs 3 step supervision learning for learning this observation to joints angle mapping, which are described below:

\begin{itemize}
\item Time contrastive supervision learning: The time contrastive (TC) supervision is used for learning a view point invariant embedded feature representation of the observations made by the learning agent.  The view point invariance helps the system to learn the action being performed by the observer independent of the view point of observation. TC supervision employs a triplet loss~\cite{DBLP:journals/corr/SchroffKP15} that brings together the similar actions from different viewpoints in the embedded space.

\item Human super vision learning: Human supervision learning is for learning the mapping between the third person observations of a human demonstrator performing an action and the joint configuration of the robot while performing similar actions. This helps the system to understand a very important mapping which is crucial for the task of observation learning, i.e the mapping between human motions and robotic motion. This mapping is very important because of the difference in the structural configurations of the human body and the robotic body. Fig ~\ref{fig:app1.3} shows this characteristics.
\begin{figure}[H]
\centering
\includegraphics[width=.3\linewidth]{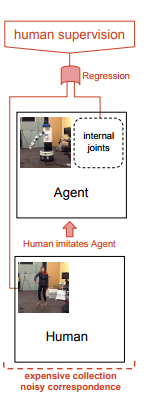}
\caption{Human supervision learning: Learning the mapping between the third person view of the demonstrator and the joint configurations}
\label{fig:app1.3}
\end{figure}

\item Self-super vision learning: The self-supervision learning is used for learning the mapping between joint angles of the robotic manipulator and the 3rd person view of itself. This helps the system to learn the direct correspondence between the joint angle movements the 3rd person observations of itself performing this action. The Fig ~\ref{fig:app1.4} demonstrates this supervision learning process.
\begin{figure}[H]
\centering
\includegraphics[width=.3\linewidth]{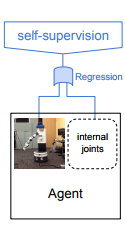}
\caption{ Self supervision learning: Learning the mapping between the third person view of the robot and the joint configurations}
\label{fig:app1.4}
\end{figure}
\end{itemize}

The net training process combines these 3 supervision learnings alternatively to learn to create an efficient mapping between observations of the demonstrator made by the learning agent to the  joint angles of the learning agent.

This works comparatively better than the previous methods in real world conditions in much less constrained environments. However training process for this method is rather difficult. It requires video samples for training in which the human demonstrator has to imitate the actions performed by the robotic manipulator so that the system can learn a human joint-robotic manipulator joint mapping. It is practically difficult to collect such samples as the human needs to imitate the actions performed by the robot exactly in the same. This exact imitation of a robot movement is time consuming and very tedious.This limits the applicability of this method.

\subsection*{Assisted observation learning}
\hspace{5ex} In assisted observation learning, the observation learning is performed by tracking a set of predefined features of the demonstrator while performing the demonstration process. The advantage of this class of methods is that the learning algorithm need not be intelligent enough to extract features by itself, as features to be extracted are hand engineered by the programmer. This makes the task much simpler. The tracked features could be hand movements of the demonstrator or predefined markers ~\cite{hamabe2015programming}~\cite{gupta2016learning}~\cite{shon2006learning}.

In the method proposed by~\citet{hamabe2015programming}, assisted observation learning is performed by tracking the hand movements of the demonstrator using the skeleton data obtained from a Kinect RGBD camera. This motion is then further used to divide the demonstrated task into simpler components and is used for learning the task. In the method by \citet{gupta2016learning} a pair of LED's are used for tracking the motions in demonstrations. Learning is then performed based on the trajectories tracked using these LED markers. In \cite{shon2006learning} pose detection is employed and the tracked feature are the poses of the demonstrator. These poses are then used to find a mapping between the action performed by the demonstrator and the joint configurations of the robotic manipulator system.

The downside of this category of observation learning methods is that, it is not always possible to extract hand engineered features for each set of tasks. The features extracted for a set of tasks need not work on another set of tasks and vice versa. More over humans do not perform observation learning by tracking a set of predefine features. Human cognition system is powerful enough to automatically learn features from raw visuals of the demonstrator performing demonstrations.

\section*{Conclusion}
\hspace{5ex} This article explains in detail the problem of observation learning along with its mathematical formulation and literature survey. It is one of the important puzzles in robotics that has to be solved for creating the next generation of consumer robotics systems. Equipping the robots with observation learning methods will make them capable of learning new skills by only visually observing a demonstrator. Even though researchers have been trying to address this problem for the past several years and are able to come up with several ways of tackling it, the field is still in its infancy and has a lot of room for development, especially in the implicit observation learning branch of observation learning problem. More advanced algorithms and methods have to be developed for giving robotic systems human like learning capability to learn new skills/tasks.

\begin{appendix}
\end{appendix}

\bibliographystyle{unsrtnat}

\end{document}